\documentclass[runningheads]{llncs}
\usepackage{array}
\usepackage[T1]{fontenc}
\usepackage{tcolorbox}
\tcbuselibrary{listings,breakable}
\usepackage{graphicx}
\usepackage{amsmath}
\usepackage[utf8]{inputenc}
\usepackage{orcidlink}

\newcounter{listing}
\renewcommand{\thelisting}{\thesection.\arabic{listing}}

\usepackage{color}

\begin{document}
\title{MinionsLLM: a Task-adaptive Framework For The
Training and Control of Multi-Agent Systems Through Natural
Language}
\titlerunning{MinionsLLM}
% If the paper title is too long for the running head, you can set
% an abbreviated paper title here

	\author{Andres Garcia Rincon\inst{1}\orcidlink{0009-0005-6121-6950}  \and
		Eliseo Ferrante\inst{1,2}\orcidlink{0000-0002-2213-8356}}
\authorrunning{A. Garcia Rincon and Ferrante}
%
% \author{Andres Garcia Rincon\inst{1}\orcidlink{0009-0005-6121-6950}}
% \author{Eliseo Ferrante\inst{1}\orcidlink{0000-0002-2213-8356}}
%

% First names are abbreviated in the running head.
% If there are more than two authors, 'et al.' is used.
%
\institute{Vrije Universiteit Amsterdam, The Netherlands \email{a.a.garciarincon@student.vu.nl, e.ferrante@vu.nl}\and
NYU Abu Dhabi, United Arab Emirates\\}

\maketitle              % typeset the header of the contribution
\begin{abstract}
This paper presents \textit{MinionsLLM}, a novel framework that integrates Large Language Models (LLMs) with Behavior Trees (BTs) and Formal Grammars to enable natural language control of multi-agent systems within arbitrary, user-defined environments. \textit{MinionsLLM} provides standardized interfaces for defining environments, agents, and behavioral primitives, and introduces two synthetic dataset generation methods (Method A and Method B) to fine-tune LLMs for improved syntactic validity and semantic task relevance. We validate our approach using Google’s Gemma 3 model family at three parameter scales (1B, 4B, and 12B) and demonstrate substantial gains: Method B increases syntactic validity to 92.6\% and achieves a mean task performance improvement of 33\% over baseline. Notably, our experiments show that smaller models benefit most from fine-tuning, suggesting promising directions for deploying compact, locally hosted LLMs in resource-constrained multi-agent control scenarios. The framework and all resources are released open-source to support reproducibility and future research.

\end{abstract}
\section{Introduction}
In nature, collective behaviors have proved effective in increasing the capabilities of individuals, from homogeneous schools of fish swarming to protect from predators, to heterogenous ant colonies with specialized roles such as workers and fighters \cite{reynolds1987flocks}. The field of swarm robotics engages in the recreation of such capabilities but faces a major challenge: while we can observe the emergent swarm behavior, it is unclear which rules the individual should follow so that it successfully emerges \cite{hasselmann2021empirical}. Some methods describe a set of primitive behaviors for the individual, such as actions to interact with the environment or conditional checks for environmental or internal conditions, and evolve a combined representation of them to achieve collective behaviors that adhere to syntactic rules defined through Formal Grammars \cite{10.1145/2463372.2463385}. Other methods define the action space for the agent and the reward structure for the desired task to induce the emergence of the desired behavior through repeated interaction with the environment \cite{Sun2024ApplicationAO}. Nevertheless, the exploration of combining Formal Grammars (FGs) and repeated environment interaction for the control of robotic agents for environment-specific tasks remains an open field.

The translation of a desired task as described by a human operator to the low-level implementation presents a significant challenge in itself. Natural Language Processing (NLP) has long been studied as a tool to overcome it. Different frameworks have achieved different levels of success by using N-tuples, a form of Agent Communication Language using shared protocols with content expressing actions or intentions\cite{Trott2015Natural}, or by using a bidirectional natural language communication pertinent to the robot and the tasks at hand with the user using task-specific controlled natural languages\cite{icaart17}. Nevertheless, these frameworks rely on predetermined communication structures that require operators to be informed and trained on them to use them, defeating the goal of facilitating swarm control for inexperienced users through Natural Language.

Large Language Models (LLMs) have emerged as one of the strongest tools in Human-Robot Interaction (HRI), demonstrating great potential in processing natural language tasks into actionable commands for agents and robots. Some frameworks focus on the direct synthesis of controllers through code generation by the LLMs \cite{strobel2024llm2swarm}, but this approach results in a framework that depends on a specific coding language, language version, and even library releases, as a simple change in the syntax of any of these components would dramatically affect the ability of the LLM to synthetize such controllers succesfully. Other frameworks intelligently employ the usage of more abstract representations for robotic controllers, like Behavior Trees \cite{zhou2024llm}. However, such frameworks heavily rely on the usage of traditional LLM service providers like OpenAI for continuous on-task prompting, which creates a dependency on massive, state-of-the-art models for API calls and continuous access to the internet. Therefore, a field of increasing interest is in enabling the deployment of such frameworks through open-source, smaller models that can be run locally with consumer-accessible hardware.

Additionally, LLMs require a degree of training in order to have a reliable task-specific performance. Frameworks like \cite{izzo2024btgenbot} choose to collect and hand-curate examples to build a general dataset, but this leads back to the problem of task-specific controller development by relying on experts' discretion, who might not know a priori a controller implementation for such a task. Other frameworks instead choose to create a semi-synthetic dataset of BTs \cite{lykov2023llm}, while very promising, the entire system design remains task-specific, and evaluation relied on several experts for different aspects of task solving. Therefore, an ongoing challenge in the field of LLMs for robot control is finding or generating datasets relevant to the user's use case, or the submersion of agents in an arbitrary environment where an LLM can use as a learning playground for training or dataset generation.

\section{The MinionsLLM Framework}
Therefore, the contribution of our paper is twofold. First, we present \textit{MinionsLLM}, to the best of our knowledge, the first LLM-based framework that enables the training and control of swarm systems for arbitrary sets of environments and agents solely through configurational changes. Second, a novel grammar-based method for dataset generation tailored to the configured agent and environment. Our framework aims to achieve this by addressing several challenges that relate to the following concepts, which will be explained in further detail in the Methodology section:

\begin{figure}[htb!]
\centering
\includegraphics[width=1\textwidth]{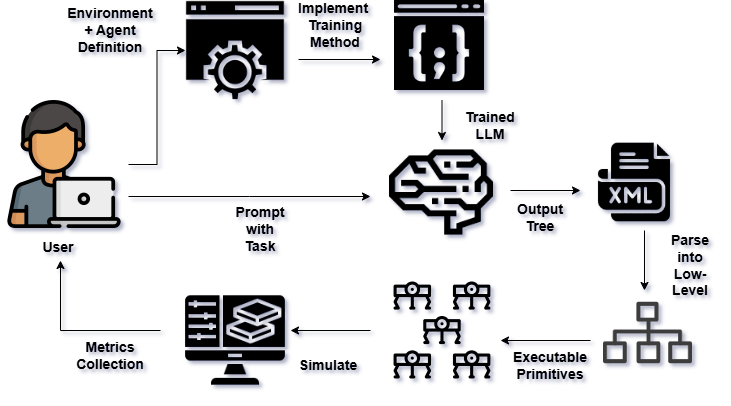}
\caption{MinionsLLM Framework Overview. User setups framework following top path. User follows bottom path to improve training method with simulation feedback} \label{datasets}
\end{figure}

\subsubsection{Standardized Agent and Environment Definition:} The \textit{MinionsLLM} framework provides a standardized way to define the agent and environment for the user's use case, enabling the streamlined setup of our framework according to the requirements and limitations of the task at hand, as explained in \ref{Agent_control}

\subsubsection{Agent Control:} Behavior Trees (BTs), further explained in Section \ref{METHOD_BT}, enable the control of agents programmatically according to the conditions the agent is subjected to. By harnessing BTs, the \textit{MinionsLLM} framework is able to abstract the low-level control of agents and robots from the LLM, which achieves two valuable feats. First, the LLM can focus on learning the proper usage of primitives instead of wasting capabilities on their implementation. Second, our framework can control any arbitrary agent, since the nodes that define the primitives can connect to anything from agents in a simulated environment, to even physical robots through frameworks like ROS \cite{koubaa2017robot}.

\subsubsection{Syntactic Structure:} For commands to be executable by an agent, they need to respect the syntax of the language used for communication. Formal Grammars (FGs) have historically provided a way to define the production rules of a language to form valid structures. The \textit{MinionsLLM} framework defines a standardized and dynamic way to define a FG to enforce the syntax of the BTs used to control such an agent to achieve tasks within such an environment. We further explain FGs in Section \ref{METHOD_FG}

\subsubsection{Semantic Relevance:} For any task to be achieved, the BTs not only need to be syntactically valid, but the combination of the agent primitive behaviors defined as nodes in the BT should form a structure with meaning within the environment. When related to language, a sentence with valid syntax but no semantics, while being parsable, would make no sense. A famous example is \textit{"Colorless green ideas sleep furiously."}. In the same way, something can not be colorless and green simultaneously; certain primitives should not be observed in certain combinations to achieve a task, even if allowed by the FG. 

This paper explores two methods to produce environment-specific synthetic datasets to finetune LLMs for the control of robotic agents, given their primitives and a FG to produce syntactically valid and semantically relevant examples of behavior trees and natural language task pairs. Nevertheless, the methodology section discusses how these methods can be easily modified to allow for methods comparable to Reinforcement Learning with Human Feedback, where task metrics and adherence to the FG take the place of the Human.

\subsubsection{Simulations as Playground for Learning:} Thanks to the abstraction of the low-level implementation of primitive behaviors, and standardized definitions for robotic agents and environments and automatic metrics collection, the \textit{MinionsLLM} Framework allows for streamlined implementation and testing of different methods to tune LLMs for Robotic control without the need for exposure to real environments.

Therefore, the \textit{MinionsLLM} framework positions itself as a playground for LLMs to train arbitrary, user-defined tasks by allowing users to use simulation metrics as direct feedback for different methods of LLM tuning. Furthermore, the \textit{MinionsLLM} framework is equipped with one such method as an example for users.

\subsubsection{Open Source:}
To contribute to the advancement of this domain, the framework will remain open source, including the dataset generation methods, and is available at: \url{https://github.com/andresgr96/MinionsLLM}.

\section{Methodology}

Several core definitions are explained before diving into the packages comprising the \textit{MinionsLLM} framework.

\subsection{Behavior Trees}
\label{METHOD_BT}
From gaming to robotics, Behavior Trees provide a modular representation for controlling agents that abstracts the low-level implementation of its nodes, which adds the benefit of remaining understandable by inexperienced users without the need for extensive explanations \cite{Ghzouli}. This structured representation allows for defining the primitives of the agents as specific types of nodes, as can be seen in Listing \ref{lst:example_bt}, according to the base nodes defined by the user and the primitive behaviors the agent is capable of performing.

% \begin{figure}[htb!]
% \centering
% \includegraphics[width=0.9\textwidth]{imgs/bt_example.jpg}
% \caption{Example of an XML Behavior Tree with Agent Primitives} \label{bt_example}
% \end{figure}

\refstepcounter{listing}
\begin{tcolorbox}[
    title={Listing~\thelisting: Example Behavior Tree},
    label={lst:example_bt},
    colback=gray!10,
    colframe=black!50!white
]
\begin{verbatim}
<BehaviorTree>
    <Selector>

        <Sequence>
            <Condition>is_good_part_detected</Condition>
            <ActuatorAction>pick_up_part</ActuatorAction>
        </Sequence>       

        <StateAction>state_seek_source_area</StateAction>

    </Selector>
</BehaviorTree>
\end{verbatim}
\end{tcolorbox}

\subsubsection{Base Node Types}
Behavior trees are constructed with nodes that dictate the logical flow of the control algorithm. Our framework defines several as base nodes, while giving the user the flexibility to add, remove, or modify existing nodes in a way they see fit for their use case by inheriting from the base class as observed in Listing \ref{lst:base_node}:

\begin{itemize}
  \item \textbf{Selector:} Executes child nodes in sequence from top to bottom until one succeeds, providing a fail-safe mechanism for attempting multiple strategies to accomplish a task.
  \item \textbf{Sequence:} Executes child nodes in sequence from top to bottom, halting if a node fails. This ensures that multi-step processes are completed correctly before executing other primitives.
  \item \textbf{Condition:} Checks specific conditions within the robot or environment, returning true or false based on real-time environmental or agent data.
  \item \textbf{Actuator Action:} Agent actions that control the actuators of the robots to interact with the environment, like picking up an object.
  \item \textbf{State Action:} Actions that change the internal state of the agent and do not necessarily affect the environment directly.
\end{itemize}

\refstepcounter{listing}
\begin{tcolorbox}[
    title={Listing~\thelisting: Base Node Abstract Class},
    label={lst:base_node},
    colback=gray!10,
    colframe=black!50!white
]
\begin{verbatim}
class BaseNode(ABC):
    """Abstract base class for all behavior tree nodes."""
    
    @abstractmethod
    def run(self, agent: Agent) -> bool:
        """
        Execute the node's behavior.

        Args:
            agent: The agent executing this node
        Returns:
            bool: True if the node succeeds, False if it fails
        Raises:
            Exception: May raise exceptions for error 
            conditions
        """
        pass
\end{verbatim}
\end{tcolorbox}

\subsection{Formal Grammars}
\label{METHOD_FG}
Formal grammars are sets of rules used to describe and generate strings within a formal language. These rules specify how to combine symbols (letters, words, etc.) to form valid "sentences" or structures within that language \cite{vauquois:hal-04701802}.
The left-hand side contains one non-terminal symbol. The right-hand side contains one or more terminal or non-terminal symbols that can be either concatenated as a logical AND or separated as options as a logical OR. The difference between terminal and non-terminal symbols is that non-terminal symbols are further expanded by a production rule, whereas terminal symbols are not.

The \textit{MinionsLLM} framework defines a formal grammar as a dictionary, as observed in Listing \ref{lst:formal_grammar}, where each key is a left-hand, non-terminal symbol and its respective values are lists of right-hand symbols that define the production rules for its key. An important notice is that all terminal symbols in our grammars have to be defined as one of the base node types that the user defines for their use case.

\refstepcounter{listing}
\begin{tcolorbox}[
    title={Listing~\thelisting: Standardized Formal Grammar Definition},
    label={lst:formal_grammar},
    colback=gray!10,
    colframe=black!50!white
]
\begin{verbatim}
grammar_rules = {                                                                 
    "B":   [["b", ["SEL"]], ["b", ["SEQ"]]],                                                          
    "SEL": [["sel", ["SEQn", "As"]], ["sel", ["SEQn"]]],                                               
    "SEQn":[["SEQ", "SEQn"], ["SEQ"]], 
    "SEQ": [["seq", ["Pn", "A"]], ["seq", ["As", "Pn", "A"]]],
    "b":   ["BehaviorTree", ["children_nodes"]],     
    "sel": ["Selector", ["children_nodes"]],
    "seq": ["Sequence", ["children_nodes"]],                                            
    "A":   [["aa", "sa"], ["aa"], ["sa"]],                                                                  
    "As":  [["aa"], ["sa"]],                                                                  
    "aa":  ["ActuatorAction"],                                                    
    "sa":  ["StateAction"],
    "Pn":  [["p", "Pn"], ["p"], []], 
    "p":   ["Condition"]
}
\end{verbatim}
\end{tcolorbox}

The first level of square brackets for a key represents a logical OR, meaning that when the symbol in the key appears in a production rule, it can be expanded into any of the lists found inside its value list. The second internal parenthesis inside each of the options for expansion is interpreted as a logical AND, where each of the symbols inside it must be expanded before that production rule is considered fully processed. Keys ending with a lowercase \textit{"n"} are interpreted as list expansions and allow for the inclusion of multiple back-to-back nodes in a row. To facilitate the understanding of the interpretation of production rules and our definition, we explain some key examples below:

\begin{itemize}
\item \textbf{"B" Rule:} an example of a symbol for which right-hand symbols are both terminal and non-terminals. When the symbol "\textit{B}" is found, two options are available. It can be expanded into the "\textit{b}" symbol, followed by the "\textit{SEL}" symbol, or into the "\textit{b}" symbol, followed by the "\textit{SEQ}" symbol. The "\textit{b}" eventually expands to the terminal "\textit{BehaviorTree}", forcing this node type as the root of all trees in our language, while the "\textit{SEQ}" and "\textit{SEL}" expand into the production options of the terminals "\textit{Sequence}" and "\textit{Selector}" respectively, forcing the root node to always contain as a child a single node, and of these two types..

\item \textbf{"p" Rule:} an example of a symbol for which right-hand symbols are only terminals. In this case, the symbol "\textit{p}" can only be expanded to the terminal symbol "\textit{Condition}", one of the base node types. 

\item \textbf{"SEQn" Rule:} an example of a list expansion rule for the symbol "\textit{SEQ}", where the first option expands to a single "\textit{SEQ}" node followed by a recursive expansion into the "\textit{SEQn}" node, allowing trees with multiple "Sequence" nodes back to back. An important observation is that different from the "\textit{Pn}" rule, which has an empty list in its third option for expansion, once this symbol is encountered during production, it is not allowed to have a list of size 0 for this node.

\end{itemize}

\subsection{Prompting Styles}
\label{prompt_styles}

Not all users share the same style of natural language communication, let alone prompting. To provide different form examples to the LLM, we define styles of prompting:

\begin{itemize}
  \item \textbf{Layman:} The most common way the average user would phrase a task. Example: "Your task is to find a good part."
  \item \textbf{Technical:} How a user with a technical background may phrase it, using logical language. 
  Example: "Your task is to find a good part, and if you do pick it up,"
  \item \textbf{Spoonfed:} Same as technical but more verbose. 
  Example: "Your task is to find a good part. If you detect a good part, pick it up."
\end{itemize}

\refstepcounter{listing}
\begin{tcolorbox}[
    title={Listing~\thelisting: Modular Prompt Format Class},
    label={lst:modular_prompt},
    colback=gray!10,
    colframe=black!50!white
]
\begin{verbatim}
{SYSTEM_PROMPT}

USER REQUEST: Generate behavior tree to "Find any part in the
environment and pick it up". Output only the XML behavior tree
without extra text or explanations of the tree.

RESPONSE: {BT_1}

USER REQUEST: Generate behavior tree to "Find all the good 
parts in the environment. If you find a good part, go to the 
base. If you are in the base then drop it there". Output 
only the XML behavior tree without extra text or explanations 
of the tree.

RESPONSE: {BT_2}

USER REQUEST: Generate behavior tree to "{PROMPT}". Output
only the XML behavior tree without extra text or explanations
of the tree.

RESPONSE: 
\end{verbatim}
\end{tcolorbox}

\subsection{Packages of the MinionsLLM Framework}

The \textit{MinionsLLM} framework is implemented as a Python package with 4 main subpackages that provide the user with different functionality.

\begin{itemize}
  \item \textbf{Interface:} This Package provides functionality to prompt LLMs locally using different backend options in order to get behavior trees.
  \item \textbf{Agent Control:} Provides functionality for the control of agents in a simulated environment through behavior trees.
  \item \textbf{Tree Parser:} Acting as middleware between the Interface and Control packages, it provides functionality to parse trees into executable commands for the agents.
  \item \textbf{Dataset Grammar:} Provides functionality to generate custom datasets through different methods.
\end{itemize}

In this section, we will go over each one of them in detail to build a hollistic view of the framework, diving deeper into the dataset generation, which remains at the core of our method and thus requires special attention.

\subsubsection{Interface}
The interface package's main functionality is to provide the user with different ways to prompt local models with tasks and receive as output a behaviour tree.

\paragraph{\textbf{Local Model Serving}}
In order to serve models locally, the \textit{MinionsLLM} framework supports two of the most popular libraries for this purpose, Ollama and LLamaCPP, providing two alternatives that increase user accessibility. Our framework utilizes the GGUF file model format specialized for running local models, and allows for both directly providing a GGUF file path, or to input a HuggingFace model URL to automatically download and set up the model for prompting.

\paragraph{\textbf{Prompt Engineering}}
Our framework provides simplified access to several prompt engineering techniques to users. By simply changing the provided template for the desired technique, the user can choose to prompt the model without instructions, add custom system prompts, or even utilize N-shot prompting with customized examples for its use case. Additionally, the primitives available to the LLM for task solving are automatically extracted from the user-specified class.

N-shot Prompting is a technique where the LLM is provided with N pairs of Prompt => Output before the actual user prompt, with the goal of informing the model of the expected output for a similar task, which has been shown to increase the performance of models over a wide array of instructions \cite{dai2022promptagator}.

The \textit{MinionsLLM} framework provides a modular format to facilitate the implementation and experimentation of different prompt engineering techniques. As can be observed in Listing \ref{lst:modular_prompt} for 2-shot prompting, the user can define the system prompt and trees to use as examples, and the framework's prompt-building process will automatically build the message in the right format for the LLM.

\subsubsection{Agent Control}
\label{Agent_control}
This package provides the user with functionality to define a simulated environment and run the generated behavior trees in a simulation.

\paragraph{\textbf{Standardized Agent Definition}}
The \textit{MinionsLLM} framework defines a standardized way to define an agent and its primitives as pythonic methods directly related to the primitives an agent can take in the environment it is subjected to. Therefore, as long as the user conforms to this standard, the framework can be used with real or simulated agents. This is thanks to the abstraction of the low-level implementations of such primitives from the LLM, which only sees their BT counterparts.

\refstepcounter{listing}
\begin{tcolorbox}[
    title={Listing~\thelisting: Agent and Primitive Definition Example},
    label={lst:agent_primitive},
    colback=gray!10,
    colframe=black!50!white
]
\begin{verbatim}
class RobotAgent(Agent):
    """
    Simplified robot agent class for example. Demonstrates 
    behavior tree integration with a basic primitive.
    """
    
    def __init__(self):
        super().__init__()
        self.is_agent_in_base_flag = True

    def is_agent_in_base_area(self) -> bool:
        """
        Node Type: Condition
        Description: Checks whether the agent is in the base
        area. Returns True if the agent is within the base, 
        and False otherwise.
        """
        return self.is_agent_in_base_flag
\end{verbatim}
\end{tcolorbox}

To define an Agent class, the user must inherit from the base \textit{"Agent"} class of the default simulator of our framework, which allows for the direct inheritance of all methods regarding Agent update in the simulation, facilitating the declaration of a new type of agent for the user. The Agent primitives can then be defined as simple class methods that evaluate to True or False, as can be observed in Listing \ref{lst:agent_primitive}, which also demonstrates the abstraction power of our framework, since the primitive simply checks for a flag. But this flag can be manipulated in several ways, depending on the use case, like the agent itself, a simulated environment, and even real-life conditions through ROS service nodes.

Finally, the user should adhere to the format observed for the documentation string of each primitive, declaring the Node Type and its description, which will be automatically extracted by our framework for Syntax checks, running primitives in the environments, and even prompt engineering.

\refstepcounter{listing}
\begin{tcolorbox}[
    title={Listing~\thelisting: Base Environment Definition},
    label={lst:base_environment},
    colback=gray!10,
    colframe=black!50!white
]
\begin{verbatim}
class SimEnvironment(ABC):
    """
    Abstract base class for all simulation environments.
    """
    
    def __init__(self, config: Config, bt_path: str):
        self.config = config
        self.simulation: Simulation = Simulation(config)
        self.xml_path = bt_path
        self.headless = headless

    @abstractmethod
    def setup(self) -> None:
        """
        Set up the environment, spawn agents, obstacles, etc.
        """
        pass

    @abstractmethod
    def run(self) -> Dict[str, Any]:
        """
        Run the simulation and return metrics/results.
        
        Returns:
            Dict[str, Any]: Dictionary containing simulation
            results and metrics
        """
        pass
\end{verbatim}
\end{tcolorbox}

The focus of our framework nevertheless remains on simulations, since they allow for the rapid testing of BTs and metric collection, which serve multiple purposes in our framework and research. \textit{MinionsLLM} provides a base environment class that all custom environments should inherit from, and forces the implementation of two functions. First is the setup function, which the user is free to implement as they see fit. It's responsible for defining environment areas, initial positioning of agents, and any other environmental aspect. 
Second, the run function, which should always run the simulation process to collect the metrics, and return them in a dictionary following the format observed in the example in Listing \ref{lst:base_environment}. Therefore, the \textit{MinionsLLM} framework allows the user to implement any arbitrary environment and calculate experimental results, filter trees by metrics achieved, and even aid in environment-specific dataset generation, as we will explain in later sections. MinionsLLM uses the Violet Simulator for all simulations as explained in Appendix \ref{simulator}.

\subsubsection{Tree Parser}
This package provides the user with functionality to parse the LLM-generated BTs into actionable commands for the agents, acting as middleware between the Interface and Control packages. 

\paragraph{\textbf{Base Node Functionality}}
Here, the user can follow the abstract class provided to implement the logical functionality for any custom node they desire. As standard, the functionality of all the nodes explained in the (BTs) section is already implemented, making sure that, for example, the Selector node follows the if-else logic expected.

\paragraph{\textbf{Tree Validation}}
Another core feature of \textit{MinionsLLM} is the ability to validate that the generated trees conform to the formal grammar. This process not only checks if the tree structure is a valid construct that follows the production rules of the grammar, but also that the primitives chosen by the LLM to solve the task are actually available to the agent class.

By allowing the user to specify both the custom grammar and the agent class, the \textit{MinionsLLM} framework can dynamically adapt to validate trees for any arbitrary use case as long as all definitions follow their specific standards.

\subsubsection{Dataset Grammar}
Thanks to the ability for users to declare custom environments, agents, and grammars, the \textit{MinionsLLM} framework enables the streamlined implementation of different mechanisms to generate datasets for any arbitrary use case. As a starting point and example for users, this package provides a mechanism to produce a fine-tuning dataset, offering different levels of control over its contents while enforcing syntactic validity. We developed two methods, hereinafter Method A and Method B, to address the challenges of generating a dataset for arbitrary environments: syntactic and semantic validity. We now endeavor to explain the parts of the general process.

\begin{figure}[b!]
\centering
\includegraphics[width=1\textwidth]{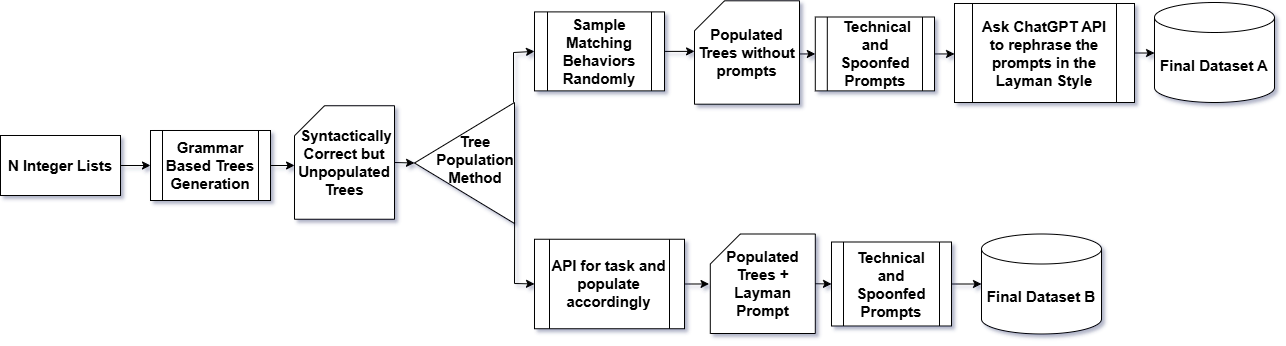}
\caption{Dataset Generation Procedure, Top and Bottom Paths Follow Methods A and B Respectively} \label{datasets}
\end{figure}

After having defined a formal grammar, agent, and environment as explained in the preceding sections, the first step is to generate a dataset of syntactically valid tree structures void of actual primitives, instead having placeholder names. For this, we adapt the method used by the authors in \cite{10.1145/2463372.2463385} to generate genotypes by iterating over a list of integers to traverse the grammar production rules. First, we randomly generate a list of \( N \) integers, where \( N_i \leq 9 \) and \( N = 10 \). Then, starting at index \( N_0 \) and with the first rule, which is always ``\textit{B}'' as shown in Listing ~\ref{lst:formal_grammar}, we count the number of options \( O \) available for that rule (in this case, \( O = 2 \)) and apply the modulo operator: \( N_0 \bmod O \).

By definition, the result of \( N_0 \bmod O \) lies in the range \( 0 \leq N_0 \bmod O < O \), so this value is a valid index into the list of options for that rule. We then select the corresponding option and expand its nonterminal symbols using the next integer \( N_1 \), continuing this process until only terminal symbols remain.

\paragraph{\textbf{Method A}}
After having generated the tree structures, for Dataset A, we randomly sample behavioral primitives that match the type of node. For example, if the node type is a condition, we check all the conditions available to the agent and sample from them. Our framework can automatically collect the available primitives and their types if the user defines the agent following the \textit{MinionsLLM} standard format. Then, we automatically generate the technical and spoonfed prompts through a proprietary script that parses the tree sequentially, which is also provided in this package. Finally, the populated tree and the technical prompt are fed to the OpenAI API with the goal of rephrasing the task in a more natural way, following the Layman style as defined in Section \ref{prompt_styles}. This process follows the top path in Figure \ref{datasets}.

\paragraph{\textbf{Method B}}
On the other hand, for Dataset B, we don't randomly populate the tree. Instead, we take inspiration from the self-instruct method, where LLMs are asked to generate datapoints after being fed examples of Input-Output pairs \cite{wang2022self}, and feed the API with the unpopulated structure and ask for it to think of a task that could be solved with it in our environment, populated the tree accordingly, and return the tree and task description, which we use as layman style. Our framework automatically validates that the LLM did not change the tree structure, including its node types, and proceeds to generate the technical and spoonfed prompt styles. This process follows the bottom path in Figure \ref{datasets}.

\paragraph{\textbf{Enhanced Control over Tree Structures}}
An attractive feature for a stochastic tree structure generation procedure is providing a degree of parametrization. The framework thus implements several parameters that can be fed to both methods in order to achieve some control over the final structure of the trees:

\begin{itemize}
    \item \textbf{\texttt{list\_max(int)}}: For list expansions, it controls the maximum number of times that recursion is allowed to \texttt{list\_max}, forcing expansion to the single version of the symbol once this limit is reached.
    \item \textbf{\texttt{list\_always(int)}}: For list expansions, it controls the exact number of nodes to be chosen, forcing the list to always contain exactly \texttt{list\_always} items.
    \item \textbf{\texttt{only(int)}}: For any rule, it forces choosing the option at the given index, regardless of the current integer in the list. 
    \item \textbf{\texttt{exclude(list)}}: For any rule, it excludes the options at the indices contained in this list, regardless of the current integer in the list.
    \item \textbf{\texttt{parent(string)}}: For any rule, if the immediate parent is of this type, it forces choosing the option at the given index, regardless of the current integer in the list. This is a dictionary, so it can be used to force different options for different parents.
\end{itemize}

This functionality enables users to steer the overall structures to more desirable shapes for their use case without the need to overcomplicate the grammar definition. Compared to natural language, it is similar to wanting to follow English grammar classes (Formal Grammar), but limiting lessons to low-complexity exercises (trees) to build the foundation towards understanding more complicated topics (tasks). 

\paragraph{\textbf{Metrics-Based Dataset Filtering}}
Thanks to the Standardized format for defining an environment, the \textit{MinionsLLM} framework offers a method to validate a tree for a given environment and agent against user-defined metrics after it's been populated before adding it to the dataset. This process enables a plethora of possibilities for ensuring semantic relevance during dataset generation, as it could be used from the user side, to look for specific metrics in trees, or inform the LLM of such metrics and validation process to ask the LLM to provide both trees and metrics, or even build an process similar to Reinforcement Learning with Human Feedback \cite{Kaufmann2023A}, where the LLM iteratively improves on a tree using the simulated environment as playground, where metric filtering combined with the syntactic validator the Parser Package offers, act as Human feedback. Nevertheless, in this paper, we limit the filtering process to the smallest meaningful metrics the agent can achieve in our experimental environment, which is explained in Section \ref{metrics}.

\paragraph{\textbf{Dataset Enrichment}}
This package additionally offers the functionality to enrich a given dataset path with N user-defined examples of LaymanTask-Tree pairs in JSON format. This process utilizes the automatic Technical and Spoonfed prompt generation process and appends the final Prompt-Tree pairs to the dataset. In this paper, we enrich each dataset with five simple examples, none of which match the tasks defined in the experimental section, to avoid test set leakage.

\section{Experimental Setup}
\subsubsection{Research Goal and Model Choice} 
The primary goal of our experimental setup is to test the full workflow of our framework, starting from agent and environment setup, to environment-specific dataset generation. The secondary goal is to study the results of the dataset generation mechanism explained in the Methodology by finetuning LLMs with the resulting datasets and comparing their performance against their non-finetuned versions.

\paragraph{\textbf{The Model}}
In order to achieve the second goal, we have chosen Google's open-source model Gemma 3. Specifically, we use their Quantization-Aware Training variants, which quantize the model weights during training. This technique allows the models to run efficiently on consumer-grade hardware by reducing memory requirements while maintaining high accuracy. \cite{nagel2022overcoming}

For all experimental units, we will test the 1B, 4B, and 12B parameter size, with the goal of analyzing the results at different sizes as well as the possible relationship between model size and improvement from baseline after finetuning.

\subsubsection{The Environment} 

\begin{figure}[htb!]
\centering
\includegraphics[width=1.0\textwidth]{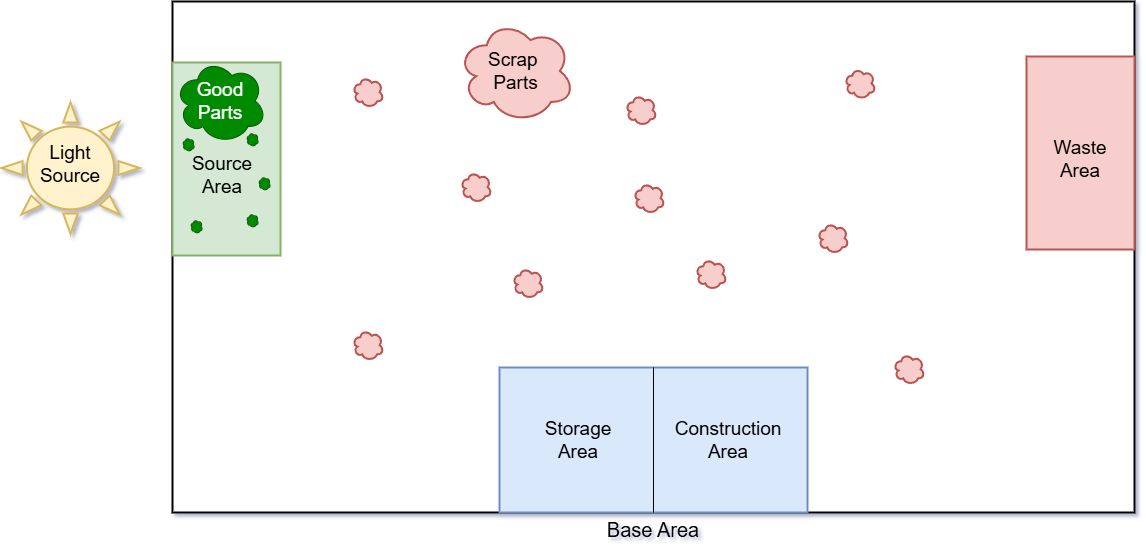}
\caption{Representation of the Base Maintenance Environment} \label{environment}
\end{figure}

For our experiments, we chose to implement a recurring theme for multi-agent systems, foraging, and base maintenance.  The agents submerged in this environment can achieve different tasks by interacting with the different elements and zones in it, as can be observed in \ref{environment}. In the environment, several parts are scattered around, and parts have two types: good parts and scrap parts. 

The environment consists of 3 main areas. The first area is the base where agents spawn and is separated into two smaller areas: a storage area and a construction area. The second area is the source area, where good parts spawn. Scrap parts can appear anywhere except the base area. The last area is the waste area, where agents can bring and drop scrap parts, so they are put away so they don't interfere with other agents. 

\subsubsection{The Agent} 

The agent is implemented as an abstract robotic agent with the only constraint of a maximum movement speed and part detection radius. It counts with a simulated sensor for the light source of the environment, allowing it to perform phototaxis to find the source, and anti-phototaxis to find the waste area, while always being able to find the direction towards its base, similar to bees.

Its primitives are thus all related to the environment. For conditions, it is able to check whether it detects or is holding any of the two kinds of parts, while also being able to check if it is within any of the environmental areas. For actuator actions, it can pick up or drop a part, while for state actions, it can change its direction of movement towards any of the areas, perform a random walk, or stop its movement altogether.

\subsubsection{Tasks and Metrics}
\label{metrics}
\begin{itemize}
  \item \textbf{Find:}  Find a good part and pick it up.
    \textbf{Metric}: Good Parts Picked up
  \item \textbf{Clean:} Find as many scrap parts as you can and bring them to the waste. 
  \textbf{Metrics}: Scrap parts picked up, Scrap parts dropped off at the waste.
  \item \textbf{Mantain:} Find as many parts as you can, bring good parts to the storage, while taking any scrap parts you find to the waste. 
  \textbf{Metrics}: Good parts picked up, Good parts dropped off at storage. Scrap parts picked up, Scrap parts dropped off at the waste.
\end{itemize}

\subsection{Core Experimental Loop}

Therefore, our experimental loop consists of the combinations shown in Table~\ref{tab:exp_loop}.

\begin{table}[h!]
    \centering
    \caption{Core Experimental Loop Configuration}
    \label{tab:exp_loop}
    \begin{tabular}{|l|l|}
        \hline
        \textbf{Category} & \textbf{Options} \\ \hline
        Prompt Techniques & Zero-shot, One-shot, Two-shot \\ \hline
        Prompt Styles & Layman, Technical, Spoon Fed \\ \hline
        Models (Instruct, 8-bit) & Gemma 3 1B, Gemma 3 4B, Gemma 3 12B \\ \hline
        Tasks & Find, Clean, Maintain \\ \hline
    \end{tabular}
\end{table}

As a baseline for our experiments, the core experimental loop is run using the vanilla models with 8B quantization. Then, we use this baseline to compare the results against the core loop, this time using the same models but finetuned with the two datasets obtained following the procedures explained in the Methodology section. The finetuning procedure is explained in Appendix Section \ref{finetuning}

\section{Results and Discussion}
\subsection{Syntactic Validity and Hallucination}

As observed in Table \ref{tab:hallucination_syntax}, both of our methods significantly increase the syntactic validity of trees produced, especially for the 1B Gemma 3 variant, with Method B proving the most effective. Additionally, while hallucination rates, defined as LLMs using primitives not available to the agent, were low for 4B and 12B, both Method A and B managed to bring it down to 0\%.

\begin{table}[h!]
    \centering
    \caption{Syntactic Validity and Hallucination Rate by Model Size}
    \label{tab:hallucination_syntax}
    \renewcommand{\arraystretch}{1.2}
    \begin{tabular}{|l|c|c|}
        \hline
        \textbf{(Baseline, A, B)} & \textbf{Syntactic Validity} & \textbf{Hallucination Rate} \\ \hline
        Gemma3\_1B & (0 - 81.5 - \textbf{88.9})\% & (NA - \textbf{0 - 0})\% \\ \hline
        Gemma3\_4B & (59.3 - 85.2 - \textbf{88.9})\% & (6.2 - \textbf{0 - 0})\% \\ \hline
        Gemma3\_12B & (59.3 - 88.9 - \textbf{100})\% & (3.1 - \textbf{0 - 0})\% \\ \hline
        \textbf{Method Mean} & (39.5 - 85.2 - \textbf{92.6})\% & (4.7 - \textbf{0 - 0})\% \\ \hline
    \end{tabular}
\end{table}

\subsection{Task Performance by Model and Prompt Technique}

As observed in Table \ref{tab:task_metrics_summary}, both of our methods significantly increase the semantic relevance of trees produced across all tasks, again observing a great increase in the smaller model size and Method B outperforming Method A. Nevertheless, the two most complex tasks, "\textit{Clean}" and "\textit{Maintain}", were only partially solved.

\begin{table}[b!]
    \centering
    \caption{Task Metrics by Model}
    \label{tab:task_metrics_summary}
    \scriptsize
    \renewcommand{\arraystretch}{1.2}
    \begin{tabular}{|l|c|c|c|c|}
        \hline
        \textbf{(Baseline, A, B)} & 
        \textbf{Good Picked Up} & 
        \textbf{Scrap Picked Up} & 
        \textbf{Good in Storage} & 
        \textbf{Scrap in Waste} \\ \hline

        find - 1B & (0 - 44.4 - \textbf{55.6})\% & NA & NA & NA \\ \hline
        find - 4B & (33.3 - 66.7 - \textbf{67.8})\% & NA & NA & NA \\ \hline
        find - 12B & (22.2 - 77.8 - \textbf{100})\% & NA & NA & NA \\ \hline
        clean - 1B & NA & (0 - 44.4 - \textbf{66.7})\% & NA & (0 - 0 - 0)\% \\ \hline
        clean - 4B & NA & (33.3 - 55.6 - \textbf{66.7})\% & NA & (0 - 0 - 0)\% \\ \hline
        clean - 12B & NA & (51.9 - 77.8 - \textbf{100})\% & NA & (0 - 0 - 0)\% \\ \hline
        maintain - 1B & (0 - 11.1 - \textbf{66.7})\% & (0 - 22.2 - \textbf{55.6})\% & (0 - 0 - 0)\% & (0 - 0 - 0)\% \\ \hline
        maintain - 4B & (0 - 11.1 - \textbf{66.7})\% & (33.3 - 22.2 - \textbf{55.6})\% & (0 - 0 - 0)\% & (0 - 0 - 0)\% \\ \hline
        maintain - 12B & (55.6 - 22.2 - \textbf{77.8})\% & (55.6 - 22.2 - \textbf{100})\% & (0 - 0 - 0)\% & (0 - 0 - 0)\% \\ \hline
    \end{tabular}
\end{table}

Additionally, an interesting finding was observed in Table \ref{tab:task_metrics_prompt}, the PE technique Zero-Shot outperformed on average One and Two-Shot. 

\begin{table}[t!]
    \centering
    \caption{Task Metrics by Prompt Technique}
    \label{tab:task_metrics_prompt}
    \scriptsize
    \renewcommand{\arraystretch}{1.2}
    \begin{tabular}{|l|c|c|c|c|}
        \hline
        \textbf{(Baseline, A, B)} & 
        \textbf{Good Picked Up} & 
        \textbf{Scrap Picked Up} & 
        \textbf{Good in Storage} & 
        \textbf{Scrap Waste} \\ \hline

        Find-Zero-Shot & (0 - \textbf{100 - 100})\% & NA & NA & NA \\ \hline
        Find-One-Shot  & (0 - 33.3 - \textbf{34.4})\% & NA & NA & NA \\ \hline
        Find-Two-Shot  & (55.6 - 55.6 - \textbf{77.8})\% & NA & NA & NA \\ \hline

        Clean-Zero-Shot & NA & (11.1 - \textbf{77.8} - 55.6)\% & NA & (0 - 0 - 0)\% \\ \hline
        Clean-One-Shot  & NA & (22.2 - 33.3 - \textbf{100})\% & NA & (0 - 0 - 0)\% \\ \hline
        Clean-Two-Shot  & NA & (51.9 - 66.7 - \textbf{77.8})\% & NA & (0 - 0 - 0)\% \\ \hline

        Maintain-Zero-Shot & (11.1 - 22.2 - \textbf{100})\% & (11.1 - 55.6 - \textbf{77.8})\% & (0 - 0 - 0)\% & (0 - 0 - 0)\% \\ \hline
        Maintain-One-Shot  & (11.1 - 11.1 - \textbf{33.3})\% & (33.3 - 33.3 - \textbf{55.6})\% & (0 - 0 - 0)\% & (0 - 0 - 0)\% \\ \hline
        Maintain-Two-Shot  & (33.3 - 11.1 - \textbf{77.8})\% & (44.4 - 33.3 - \textbf{77.8})\% & (0 - 0 - 0)\% & (0 - 0 - 0)\% \\ \hline
    \end{tabular}
\end{table}

\subsection{Mean Performance Increase By Model Size and Method}

\begin{figure}[t!]
\centering
\includegraphics[width=1.0\textwidth]{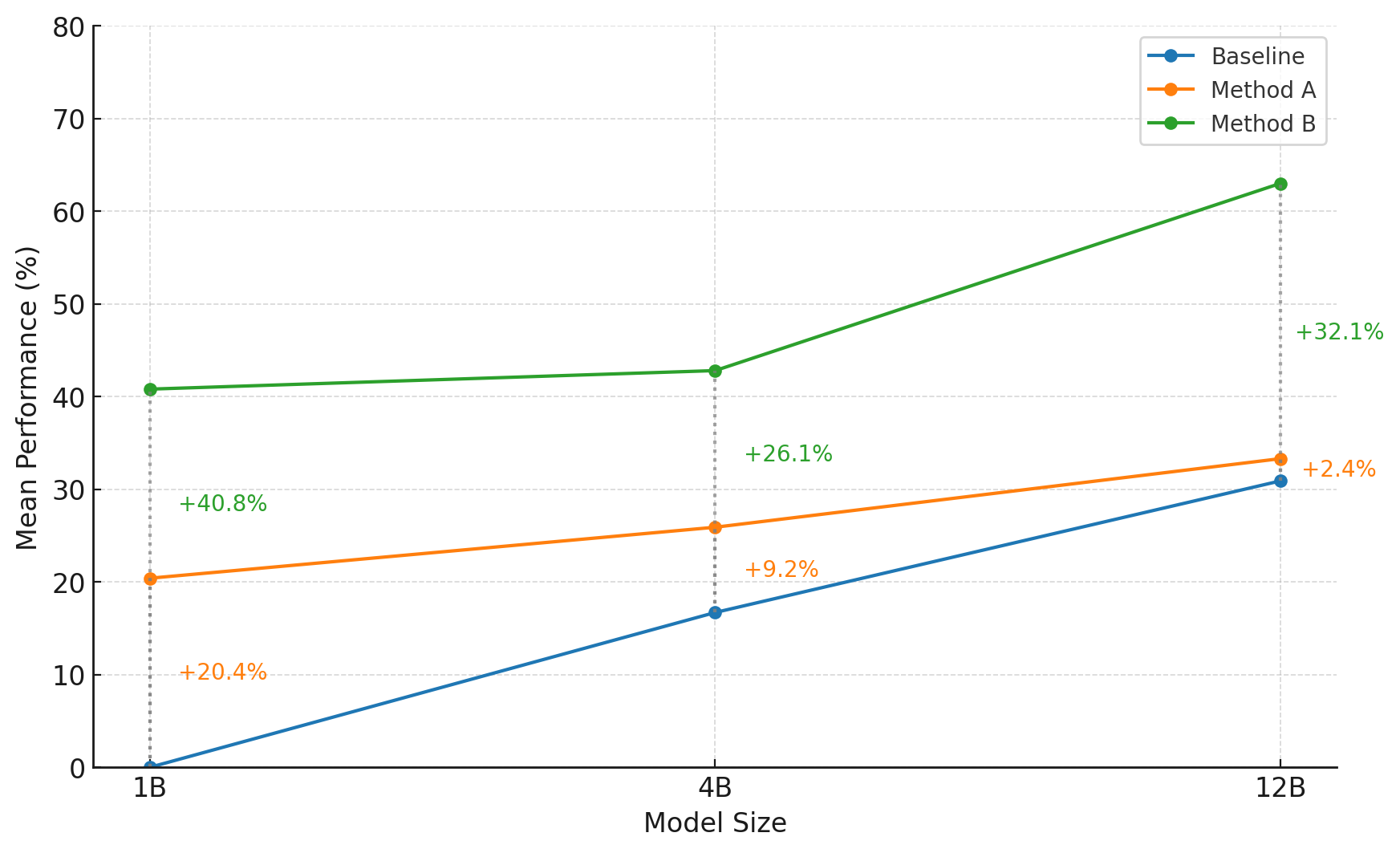}
\caption{Mean Task Performance by Model Size} \label{mean_perf}
\end{figure}

We are also interested in studying the relationship between the size of the model and the observed benefit of finetuning. As observed in Figure \ref{mean_perf}, all methods show a tendency towards a linear relationship between model size and task performance. Additionally, for both of our methods, the smallest size showed the biggest improvement in performance.

\begin{table}[b!]
    \centering
    \caption{Mean Task Performance by Model Size and Method (All Metrics)}
    \label{tab:mean_perf_by_size_all}
    \renewcommand{\arraystretch}{1.2}
    \begin{tabular}{|l|c|c|c|c|}
        \hline
        \textbf{Model Size} & 
        \textbf{Baseline} & 
        \textbf{Method A} & 
        \textbf{Method B} & 
        \textbf{Mean Gain By Size} \\ \hline

       1B & 0.0\% & 20.4\% (+20.4\%) & 40.8\% \textbf{(+40.8\%)} & \textbf{30.6\%} \\ \hline
        4B & 16.7\% & 25.9\% (+9.3\%) & 42.8\% \textbf{(+26.2\%)} & 17.7\% \\ \hline
        12B & 30.9\% & 33.3\% (+2.4\%) & 63.0\% \textbf{(+32.1\%)} & 17.3\% \\ \hline
        \textbf{Mean} & 15.8\% & 26.5\% (+10.7\%) & 48.8\% \textbf{(+33.0\%)} & NA \\ \hline
    \end{tabular}
\end{table}

\subsection{Discussion}
The main goal of our experiments was to test the complete workflow of the \textit{MinionsLLM} framework in a simulated environment to assess its capabilities and potential as a system for training and controlling agents in arbitrary settings. We identified two key challenges such frameworks must tackle. The first is \textit{Syntactic Validity}. Baseline models averaged 39.5\% syntactic validity across all sizes, with the smallest 1B model achieving only 0\%. Both Method~A and Method~B significantly improved these rates to 85.2\% and 92.6\% respectively, with the largest model under Method~B reaching 100\% validity. This demonstrates that our approach can effectively teach syntactic rules to LLMs without explicit instruction.

The second aspect is \textit{Semantic relevance}, meaning the LLM’s ability to use the agent’s primitives to solve tasks. We tested this with three tasks of increasing complexity: \textit{Find}, \textit{Clean}, and \textit{Maintain}, which require agents to pick up a part, pick up and drop a part, and handle multiple part types in different areas, respectively. Both methods improved performance, but only Method B showed a significant mean increase of 33\% (Table~\ref{tab:task_metrics_summary}). However, the second metric for \textit{Clean} and \textit{Maintain} was never achieved, indicating our methods can only teach LLMs simple task relevance so far. In linguistic terms, this is akin to a student mastering basic sentences but struggling with more complex ones, resulting in only partially meaningful output.

Additionally, an interesting finding was observed when analyzing the performance of the different N-shot techniques: in most cases, Zero-Shot outperformed both providing one and two examples as part of the prompt, as can be seen in Table \ref{tab:task_metrics_prompt}. We believe this is comparable to a student having internalized correctly the language lessons, but thanks to the recency and similarity of the examples given by the teacher prior to a test, the student adapts their answers to resemble the process the teacher just showed as correct. Nevertheless, this finding requires further study on the effect of examples on recency bias for LLMs in domains with little prior training, as our environments are arbitrary, it could be harder for LLMs to adapt examples effectively.

Finally, as observed in Table \ref{tab:mean_perf_by_size_all}, the smallest model of 1B showed the largest increase in performance, with Method B resulting in a 40.8\% performance increase across all tasks. Combined with the fact that this size also showed the largest increase in syntactic validity, these results indicate great promise in finetuning very small models to perform tasks within controlled, arbitrary environments.

\section{Conclusion}
This paper introduced \textit{MinionsLLM}, a framework that lets users define arbitrary environments and agents in simulation as a testbed for fine-tuning LLMs for multi-agent task solving with small, locally deployable models that, once trained, no longer depend on credits or internet access. We showed that \textit{MinionsLLM} streamlines research on training methods to improve both syntactic validity and semantic relevance (task performance), even with models as small as 1B and 8-bit quantization. To achieve this, we presented two methods that use formal grammars with \textit{MinionsLLM} to generate synthetic datasets tailored to the user’s needs. The best performing Method B achieved a mean syntactic validity Rate of 92.6\% and a mean task performance score of 40.8\%, an increase of 53.1\% and 33\% from baseline, respectively. Finally, our experiments found that the largest increase in performance was observed in the smallest 1B model, and the relationship between increase in task performance and model size remained negative for the most part, indicating great promise in the usage of smaller, locally deployable models for the control of Multi-Agent Systems.

\bibliographystyle{splncs04}
\bibliography{minions}

\begin{thebibliography}{10}
\providecommand{\url}[1]{\texttt{#1}}
\providecommand{\urlprefix}{URL }
\providecommand{\doi}[1]{https://doi.org/#1}

\bibitem{dai2022promptagator}
Dai, Z., Zhao, V.Y., Ma, J., Luan, Y., Ni, J., Lu, J., Bakalov, A., Guu, K., Hall, K.B., Chang, M.W.: Promptagator: Few-shot dense retrieval from 8 examples. arXiv preprint arXiv:2209.11755  (2022)

\bibitem{10.1145/2463372.2463385}
Ferrante, E., Du\'{e}\~{n}ez Guzm\'{a}n, E., Turgut, A.E., Wenseleers, T.: Geswarm: grammatical evolution for the automatic synthesis of collective behaviors in swarm robotics. In: Proceedings of the 15th Annual Conference on Genetic and Evolutionary Computation. p. 17–24. GECCO '13, Association for Computing Machinery, New York, NY, USA (2013). \doi{10.1145/2463372.2463385}, \url{https://doi.org/10.1145/2463372.2463385}

\bibitem{Ghzouli}
Ghzouli, R., Berger, T., Johnsen, E.B., Dragule, S., W\k{a}sowski, A.: Behavior trees in action: a study of robotics applications. In: Proceedings of the 13th ACM SIGPLAN International Conference on Software Language Engineering. p. 196–209. SLE 2020, Association for Computing Machinery, New York, NY, USA (2020). \doi{10.1145/3426425.3426942}, \url{https://doi.org/10.1145/3426425.3426942}

\bibitem{hasselmann2021empirical}
Hasselmann, K., Ligot, A., Ruddick, J., Birattari, M.: Empirical assessment and comparison of neuro-evolutionary methods for the automatic off-line design of robot swarms. Nature communications  \textbf{12}(1), ~4345 (2021)

\bibitem{hu2022lora}
Hu, E.J., Shen, Y., Wallis, P., Allen-Zhu, Z., Li, Y., Wang, S., Wang, L., Chen, W., et~al.: Lora: Low-rank adaptation of large language models. ICLR  \textbf{1}(2), ~3 (2022)

\bibitem{izzo2024btgenbot}
Izzo, R.A., Bardaro, G., Matteucci, M.: Btgenbot: Behavior tree generation for robotic tasks with lightweight llms. In: 2024 IEEE/RSJ International Conference on Intelligent Robots and Systems (IROS). pp. 9684--9690. IEEE (2024)

\bibitem{Kaufmann2023A}
Kaufmann, T., Weng, P., Bengs, V., Hüllermeier, E.: A survey of reinforcement learning from human feedback. ArXiv  \textbf{abs/2312.14925} (2023). \doi{10.48550/arXiv.2312.14925}

\bibitem{koubaa2017robot}
Koubaa, A., et~al.: Robot Operating System (ROS)., vol.~1. Springer (2017)

\bibitem{lykov2023llm}
Lykov, A., Dronova, M., Naglov, N., Litvinov, M., Satsevich, S., Bazhenov, A., Berman, V., Shcherbak, A., Tsetserukou, D.: Llm-mars: Large language model for behavior tree generation and nlp-enhanced dialogue in multi-agent robot systems. arXiv preprint arXiv:2312.09348  (2023)

\bibitem{icaart17}
Mészáros, T., Dobrowiecki, T.: Agent-based reconfigurable natural language interface to robots - human-agent interaction using task-specific controlled natural languages. In: Proceedings of the 9th International Conference on Agents and Artificial Intelligence - Volume 1: ICAART. pp. 632--639. INSTICC, SciTePress (2017). \doi{10.5220/0006205306320639}

\bibitem{nagel2022overcoming}
Nagel, M., Fournarakis, M., Bondarenko, Y., Blankevoort, T.: Overcoming oscillations in quantization-aware training. In: International Conference on Machine Learning. pp. 16318--16330. PMLR (2022)

\bibitem{reynolds1987flocks}
Reynolds, C.W.: Flocks, herds and schools: A distributed behavioral model. In: Proceedings of the 14th annual conference on Computer graphics and interactive techniques. pp. 25--34 (1987)

\bibitem{strobel2024llm2swarm}
Strobel, V., Dorigo, M., Fritz, M.: Llm2swarm: robot swarms that responsively reason, plan, and collaborate through llms. arXiv preprint arXiv:2410.11387  (2024)

\bibitem{Sun2024ApplicationAO}
Sun, Q., Chen, Z., Liu, H.: Application and optimization of multi-agent reinforcement learning in collaborative decision-making. In: International Conference on Innovative Computing and Cloud Computing (2024), \url{https://api.semanticscholar.org/CorpusID:274657640}

\bibitem{Trott2015Natural}
Trott, S., Feldman, J., Janin, A.L., Appriou, A.: Natural language understanding and communication for multi-agent systems pp. 137--141 (2015)

\bibitem{vauquois:hal-04701802}
Vauquois, B.: {A Survey of Formal Grammars and Algorithms for Recognition and Transformation in Machine Translation}. In: {IFIP Congress-68}. No. 254-260 in Bernard Vauquois et la TAO - Analectes, Edimburg, United Kingdom (1968), \url{https://hal.science/hal-04701802}, revised and re-edited by Ch. Boitet, May 1988

\bibitem{wang2022self}
Wang, Y., Kordi, Y., Mishra, S., Liu, A., Smith, N.A., Khashabi, D., Hajishirzi, H.: Self-instruct: Aligning language models with self-generated instructions. arXiv preprint arXiv:2212.10560  (2022)

\bibitem{zhou2024llm}
Zhou, H., Lin, Y., Yan, L., Zhu, J., Min, H.: Llm-bt: Performing robotic adaptive tasks based on large language models and behavior trees. In: 2024 IEEE International Conference on Robotics and Automation (ICRA). pp. 16655--16661. IEEE (2024)

\end{thebibliography}

\newpage
% \begin{appendices}

\section*{Appendix}\label{secA1}
\subsection{Model Finetuning}
\label{finetuning}
Fine-tuning large language models across all layers is computationally expensive and time-consuming. To address this, parameter-efficient fine-tuning (PEFT) methods have emerged as effective alternatives. Although many PEFT techniques can underperform compared to full fine-tuning, Low-Rank Adaptation (LoRA) has demonstrated competitive or superior results in some cases by mitigating catastrophic forgetting—where the model loses pre-trained knowledge during adaptation  \cite{hu2022lora}.

This is achieved by freezing the original weight matrices and instead learning two low-rank matrices, known as LoRA adapters, which approximate updates to the full weight matrix. The trained adapters are then merged with the pre-trained model for inference. QLoRA further improves memory efficiency by loading the pre-trained weights in 4-bit quantized form (versus 8-bit for standard LoRA) while maintaining comparable performance. We therefore used QlLoRA in combination with \textit{Unsloth}, one of the most used Python libraries for finetuning, in order to finetune our models.
\subsubsection{Finetuning Procedure}
In order to introduce as little room for error as possible, we utilize the pre-prepared Unsloth Notebook for the fine-tuning of Gemma 3 Models available publicly at \url{https://colab.research.google.com/github/unslothai/notebooks/blob/main/nb/Gemma3_(4B).ipynb}

The notebook was modified to adapt to our model choice and custom dataset. Parameters for the trainer are as defined in Table \ref{tab:finetune_params}, with special attention to the fact that only three epochs were carried out, which, given a 300 dataset size, resulted in 111 training steps. 

\begin{table}[h!]
    \centering
    \caption{Fine-tuning Hyperparameters for SFT}
    \label{tab:finetune_params}
    \renewcommand{\arraystretch}{1.2}
    \begin{tabular}{|l|c|}
        \hline
        \textbf{Hyperparameter} & \textbf{Value} \\ \hline
        Per-device batch size & 2 \\ \hline
        Gradient accumulation steps & 4 \\ \hline
        Warmup steps & 5 \\ \hline
        Number of epochs & 3 \\ \hline
        Learning rate & 2e-4 \\ \hline
        Optimizer & AdamW (8-bit) \\ \hline
        Weight decay & 0.01 \\ \hline
        LR scheduler & Linear \\ \hline
        Logging steps & 1 \\ \hline
    \end{tabular}
\end{table}

\subsection{The Violet Simulator}
\label{simulator}

We use the Python simulator Violet available at \url{https://github.com/m-rots/violet} to build and run our simulated environments. Violet is built on top of the well-known PyGame Python library and was built with multi-agent simulation as its core goal. It offers unparalleled simulation speed through several game development-inspired methodologies for multi-agent interaction calculations, enabling MinionsLLMs to run thousands of simulations in a matter of minutes.

\begin{figure}[b!]
\centering
\includegraphics[width=1.0\textwidth]{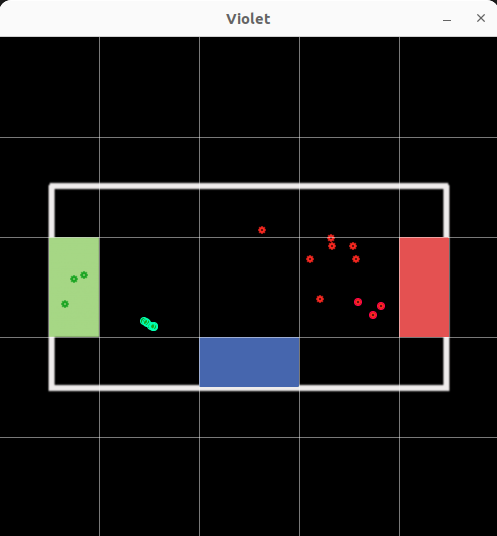}
\caption{Experimental Environment on the Violet Simulator} \label{viviviv}
\end{figure}

% \end{appendices}

\end{document}